\pdfoutput=1
\documentclass[11pt]{article}

\usepackage[preprint]{acl}

\usepackage{times}
\usepackage{latexsym}

\usepackage[T1]{fontenc}
\usepackage[utf8]{inputenc}

\usepackage{microtype}

\usepackage{inconsolata}

\usepackage{graphicx}

\title{DeepPsy-Agent: A Stage-Aware and Deep-Thinking Emotional Support Agent System}

\author{Kai Chen \\
  Research Center for \\
  Social Psychological Service\\
  Tsinghua University \\
  \texttt{chenkai2020@pku.org.cn} \\\And
  Zebing Sun \\
  Research Center for \\
  Social Psychological Service\\
  \texttt{sunzebing@h-psy.com} \\
  }

\begin{document}
\maketitle
\begin{abstract}
This paper introduces DeepPsy-Agent, an innovative psychological support system that combines the three-stage helping theory in psychology with deep learning techniques. The system consists of two core components: (1) a multi-stage response-capable dialogue model (\textit{deeppsy-chat}), which enhances reasoning capabilities through stage-awareness and deep-thinking analysis to generate high-quality responses; and (2) a real-time stage transition detection model that identifies contextual shifts to guide the dialogue towards more effective intervention stages. Based on 30,000 real psychological hotline conversations, we employ AI-simulated dialogues and expert re-annotation strategies to construct a high-quality multi-turn dialogue dataset. Experimental results demonstrate that DeepPsy-Agent outperforms general-purpose large language models (LLMs) in key metrics such as problem exposure completeness, cognitive restructuring success rate, and action adoption rate. Ablation studies further validate the effectiveness of stage-awareness and deep-thinking modules, showing that stage information contributes 42.3\% to performance, while the deep-thinking module increases root-cause identification by 58.3\% and reduces ineffective suggestions by 72.1\%. This system addresses critical challenges in AI-based psychological support through dynamic dialogue management and deep reasoning, advancing intelligent mental health services.
\end{abstract}

\section{Introduction}

Emotional Support Conversation (ESC), as a critical technical carrier for mental health services, has emerged as a research hotspot at the intersection of artificial intelligence and psychology \cite{deng2023knowledge,zhou2018emotional,li2017dailydialog}. Traditional ESC systems are predominantly rule-driven or rely on general-purpose large language models\cite{li2017dailydialog,sharma2020computational}. While these systems have made progress in emotion recognition and surface-level responses, they face bottlenecks in \textbf{multi-stage disconnection} and \textbf{shallow reasoning}. For example, Liu et al.'s multi-strategy fusion approach divides dialogue stages but lacks personalized belief modeling due to the generality of strategy annotations \cite{liu2021towards}; Li et al. enhance dialogue understanding through commonsense knowledge bases but struggle to achieve dynamic adaptation for emotional support \cite{Li2022MISC,Li2022c3kg}. Existing ESC systems often fail to simultaneously optimize \textbf{problem exploration} (e.g., identifying stressors), \textbf{cognitive restructuring} (e.g., establishing metaphorical associations), and \textbf{action guidance} (e.g., step-by-step suggestions) \cite{deng2023knowledge}, stemming from two core challenges:

\begin{enumerate}
    \item \textbf{Lack of Stage Dynamics}: Most systems adopt fixed processes, unable to adjust stage strategies in real-time based on dialogue context. This leads to premature action suggestions (occurring in 37\% of cases) or ineffective loops \cite{hill2009}.
    \item \textbf{Explainability Limitations in Reasoning}: General LLMs rely on probabilistic generation, lacking deep reasoning chains about users' emotional states and cognitive patterns. As a result, responses tend to be generic (e.g., "I believe in you") and fail to address root causes \cite{20241377}.
\end{enumerate}

The innovation of this study lies in the first integration of psychological theory frameworks with deep reasoning techniques to build an ESC system with \textbf{dynamic stage awareness} and \textbf{enhanced cognitive depth}, overcoming the static and shallow limitations of traditional methods.

Based on Professor Clara Hill's \textbf{Exploration-Insight-Action three-stage model} \cite{hill2009}, we developed a theory-driven dynamic dialogue management system:
\begin{enumerate}
    \item \textbf{Exploration Stage}: Through open-ended questions (e.g., "Can you describe a recent situation where you felt stressed?"), the system guides users to express themselves deeply. It extracts emotional keywords (e.g., "anxiety," "isolation") and semantic foci (e.g., workplace conflicts, family relationships) in real-time. This stage emphasizes \textbf{non-judgmental listening}, avoiding premature solution interventions (e.g., common advice like "talk to a friend"), increasing problem exposure completeness by 58\%.
    \item \textbf{Insight Stage}: Combining cognitive-behavioral theory, the system generates \textbf{metaphorical frameworks} (e.g., "Does this feeling of being 'trapped' resemble how you felt during your previous business failure?") and \textbf{causal reasoning chains} (e.g., "High workload → sleep deprivation → emotional breakdown"), helping users restructure their understanding of the problem. Experiments show that analogical reasoning in this stage increases cognitive restructuring success rates from 12\% to 78\%.
    \item \textbf{Action Stage}: Using \textbf{gradual suggestion generation}, the system breaks solutions into executable sub-steps (e.g., "In the first week, record daily emotional triggers; in the second week, try a 10-minute meditation session"). It validates feasibility based on user resources (e.g., time, social support), increasing action adoption rates by 2.3 times.
\end{enumerate}

The core value of \textbf{stage awareness} lies in its dynamic transition mechanism. The system uses a hierarchical classification model (with 98.2\% accuracy) to detect nine types of stage transition signals in real-time (e.g., "insight → action," "crisis handling trigger") and adjusts generation strategies accordingly. For instance, when detecting repeated avoidance of stressor descriptions, the system automatically reverts to the exploration stage, rebuilding trust through empathetic reflections (e.g., "It sounds like this matter makes you feel conflicted"), avoiding rigid linear processes in traditional systems.

To address the reasoning shortcomings of LLMs in ESC scenarios, this study trained a dialogue model (\textit{deep-psy-chat}) with deep reasoning capabilities on 30,000 psychological service dialogues based on pretrained models\cite{qwen,deepseekaiy}. Its innovations include:
\begin{enumerate}
    \item \textbf{Stage Awareness in Dialogue}: Based on the three-stage helping theory, the model dynamically perceives dialogue context using explicit stage markers (e.g., \texttt{<Exploration>}, \texttt{<Insight>}, \texttt{<Action>}), ensuring generated content aligns with current stage goals. For example, it prioritizes open-ended questions in the exploration stage and focuses on step-by-step suggestions in the action stage. The system features stage transition backtracking, reverting to the insight stage if suggestions (e.g., "try talking to your supervisor") elicit resistance. This enhances dialogue coherence and reduces ineffective suggestions.
    \item \textbf{Long-Chain Logical Reasoning}: Driven by Test-time scaling's "slow thinking" mechanism, the model implicitly integrates multi-source information before generating responses, forming reasoning paths consistent with psychological support logic.
    \item \textbf{Enhanced Explainability}: By explicitly outputting reasoning chains (e.g., "Current stage: Exploration; Focus: Stress-triggering scenarios"), the system makes the generation process transparent, improving user acceptance. Additionally, the model provides reasoning justifications at key decision points (e.g., "Based on the family pressure you mentioned, we suggest prioritizing..."), further enhancing user trust.
\end{enumerate}

Through the synergistic innovation of the \textbf{three-stage theory} and \textbf{deep reasoning model}, this study achieves a paradigm shift in ESC systems:
\begin{enumerate}
    \item \textbf{Dynamic Adaptability}: The stage-awareness mechanism ensures dialogue paths align with helping principles, reducing ineffective suggestions to 5\%, the best in the industry.
    \item \textbf{Strong Reasoning Capability}: Driven by Test-time scaling's "slow thinking" mechanism, the model implicitly integrates multi-source information before generating responses, forming reasoning paths consistent with psychological support logic.
    \item \textbf{Enhanced Explainability}: By explicitly outputting reasoning chains (e.g., "Current stage: Exploration; Focus: Stress-triggering scenarios"), the system makes the generation process transparent.
\end{enumerate}

\section{Methodology}

\subsection{System Architecture}

The overall architecture of DeepPsy-Agent is illustrated in Figure X, designed to combine psychological theories with deep learning technologies to create a psychologically supportive dialogue system with dynamic stage awareness and deep reasoning capabilities. The system comprises several key modules working together to ensure coherence, effectiveness, and quality in emotional support.

\subsubsection{Dialogue Model (\textit{deeppsy-chat})}

The dialogue model is the core component of DeepPsy-Agent, responsible for generating high-quality responses aligned with the current dialogue stage. It builds upon pre-trained large language models (LLMs) and incorporates the following mechanisms:
\begin{itemize}
    \item \textbf{Stage Awareness Mechanism}: The model dynamically perceives dialogue context using explicit stage markers (e.g., \texttt{<Exploration>}, \texttt{<Insight>}, \texttt{<Action>}) to align generated content with stage-specific goals. For example, it prioritizes open-ended questions during exploration and focuses on step-by-step advice during action phases.
    \item \textbf{Deep Thinking}: The model employs Test-time scaling's "slow thinking" mechanism to implicitly integrate multi-source information before generating responses, forming reasoning paths consistent with psychological support logic.
\end{itemize}

\subsubsection{Real-Time Stage Transition Detection Model}

This model captures stage transition signals in dialogues to provide dynamic adjustment cues for the dialogue model. Trained on 30,000 real psychological hotline conversations, it achieves 98.2\% accuracy in identifying nine types of transition signals (e.g., "insight → action", "crisis handling trigger").

\subsubsection{State Information Update}

The state information update module maintains a dynamic state repository to track historical dialogues, current states, and transition signals. Key functionalities include:
\begin{itemize}
    \item Historical dialogue tracking to store emotional keywords (e.g., "anxiety", "isolation"), semantic foci (e.g., workplace conflicts, family relationships), and potential stress triggers.
    \item Real-time updates based on user inputs and transition signals to ensure contextually relevant responses.
    \item Resource feasibility assessment during the action phase to validate suggestions against user availability (e.g., time, social support).
\end{itemize}

\subsection{Chat Model Training}

We fine-tuned the \textit{deeppsy-chat} model based on pre-trained \textit{qwq32b} and \textit{qwen14b} models. The training process is outlined below:

\subsubsection{Data Preprocessing}

To ensure data quality and diversity, we applied the following preprocessing steps:
\begin{itemize}
    \item Constructed a dataset of 30,000 real psychological hotline conversations covering typical interactions across exploration, insight, and action stages. Data were anonymized and enriched with user emotions and background information.
    \item Employed a two-phase data generation strategy:
    \begin{itemize}
        \item AI-simulated dialogues to expand dataset size and coverage.
        \item Expert re-annotation to ensure alignment with psychological theories (e.g., Clara Hill’s three-stage model) and optimize logical coherence and intervention effectiveness.
    \end{itemize}
    \item Embedded explicit stage markers (e.g., \texttt{<Exploration>}) and reasoning chains (e.g., "high workload → sleep deprivation → emotional breakdown") to facilitate stage-awareness and reasoning during training.
\end{itemize}

% \subsubsection{Pre-trained Model Initialization}

% We initialized the \textit{qwq32b} and \textit{qwen14b} models due to their strong language understanding and generation capabilities:
% \begin{itemize}
%     \item \textit{qwq32b}: Excels in open-ended questioning and emotional expression tasks.
%     \item \textit{qwen14b}: Offers higher inference efficiency, suitable for real-time dialogues.
% \end{itemize}

\subsubsection{Fine-Tuning}

During fine-tuning, the model naturally acquired stage-awareness and deep-thinking capabilities from the annotated data:
\begin{itemize}
    \item \textbf{Stage Awareness}: Generated open-ended questions during exploration and step-by-step advice during action phases.
    \item \textbf{Deep Thinking}: Integrated reasoning chains (e.g., causal reasoning, metaphorical associations) to form psychologically sound reasoning paths.
\end{itemize}

\section{Experiments}

To comprehensively evaluate DeepPsy-Agent, we conducted comparative and ablation experiments, supplemented by case studies.

\subsection{Comparative Experiments}

We compared the following four models:
\begin{enumerate}
    \item \textbf{qwen-qwq (untrained)}: A general-purpose LLM without specialized training.
    \item \textbf{qwen32B}: A larger-parameter LLM.
    \item \textbf{deepseek32B}: Another multi-domain LLM.
    \item \textbf{DeepPsy-Agent}: Fine-tuned on \textit{qwq32b} and \textit{qwen14b}, incorporating stage-awareness and deep-thinking capabilities.
\end{enumerate}

Results show that DeepPsy-Agent significantly outperformed others in key metrics:
\begin{itemize}
    \item Problem Exposure Completeness: 
    \begin{itemize}
        \item qwen-qwq: 42.1\%
        \item qwen32B: 56.8\%
        \item deepseek32B: 59.4\%
        \item DeepPsy-Agent: \textbf{78.3\%}
    \end{itemize}
    \item Cognitive Restructuring Success Rate:
    \begin{itemize}
        \item qwen-qwq: 11.2\%
        \item qwen32B: 23.7\%
        \item deepseek32B: 27.5\%
        \item DeepPsy-Agent: \textbf{78.0\%}
    \end{itemize}
    \item Action Adoption Rate:
    \begin{itemize}
        \item qwen-qwq: 15.3\%
        \item qwen32B: 31.6\%
        \item deepseek32B: 35.8\%
        \item DeepPsy-Agent: \textbf{68.9\%}
    \end{itemize}
\end{itemize}

\subsection{Ablation Studies}

We conducted two ablation experiments:
\begin{enumerate}
    \item \textbf{With vs. Without Stage Information}: Removing stage-awareness reduced:
    \begin{itemize}
        \item Problem Exposure Completeness by 42.3\%.
        \item Cognitive Restructuring Success Rate by 35.7\%.
        \item Action Adoption Rate by 27.8\%.
    \end{itemize}
    \item \textbf{With vs. Without Deep Thinking}: Removing deep-thinking reduced:
    \begin{itemize}
        \item Root-Cause Identification by 58.3\%.
        \item Increased ineffective suggestions by 72.1\%.
    \end{itemize}
\end{enumerate}

\subsection{Case Studies}

We analyzed several typical dialogue cases to highlight DeepPsy-Agent’s advantages.

\paragraph{Case 1: Non-Judgmental Listening in Exploration}
\begin{itemize}
    \item User Input: "I've been feeling overwhelmed by work lately."
    \item DeepPsy-Agent Output: "It sounds like your recent workload has been burdensome. Could you describe the specific situations causing this pressure? Is it related to tasks, colleagues, or supervisors?"
    \item qwen32B Output: "I understand how you feel. You could try talking to friends or taking some time to relax daily."
    \item Analysis: DeepPsy-Agent uses open-ended questions to encourage elaboration, avoiding premature solutions.
\end{itemize}

\paragraph{Case 2: Cognitive Restructuring in Insight}
\begin{itemize}
    \item User Input: "I feel trapped, no matter how hard I try, I see no hope."
    \item DeepPsy-Agent Output: "Does this feeling of being 'trapped' resemble how you felt during your previous business failure? Sometimes past experiences shape our future expectations negatively. Do you think there might be similar reasons for your current situation?"
    \item deepseek32B Output: "Believe in yourself; things will improve if you keep trying."
    \item Analysis: DeepPsy-Agent facilitates cognitive restructuring through analogical reasoning, whereas deepseek32B offers a generic response.
\end{itemize}

\paragraph{Case 3: Gradual Suggestions in Action}
\begin{itemize}
    \item User Input: "I know I should change my situation, but I don't know where to start."
    \item DeepPsy-Agent Output: "Let's start small. In the first week, record daily emotional triggers to identify stressful situations. In the second week, try a 10-minute meditation session to relieve anxiety."
    \item qwen-qwq Output: "I suggest creating a detailed plan and executing it step by step."
    \item Analysis: DeepPsy-Agent provides actionable, step-by-step advice, while qwen-qwq's suggestion is too vague.
\end{itemize}

\section{Conclusion}

Through comparative and ablation experiments, we validated DeepPsy-Agent's superior performance in emotional support tasks. Case studies demonstrated its effectiveness in non-judgmental listening, cognitive restructuring, and gradual action guidance, offering users deeper and more effective emotional support.

\bibliographystyle{acl_natbib} % 使用 ACL 样式
\bibliography{references}      % 引用 references.bib 文件

% \begin{thebibliography}{9}

% \bibitem{deng2023knowledge}
% Deng, W., Zhang, L. (2023). Knowledge Graphs for Emotional Support Systems. \textit{Journal of AI Research}, 15, 123--145.

% \bibitem{zhou2018emotional}
% Zhou, H., Huang, M., Zhang, T., Zhu, X., Liu, B. (2018). Emotional Chatting Machine: Emotional Conversation Generation with Internal and External Memory. In \textit{Proceedings of the AAAI Conference on Artificial Intelligence}, 32.

% \end{thebibliography}

\end{document}